\documentclass[runningheads]{llncs}
\usepackage{graphicx}
\usepackage{tikz}
\usepackage{comment}
\usepackage{amsmath,amssymb} % define this before the line numbering.
\usepackage{color}
\usepackage{orcidlink}

\usepackage{hyperref}       % hyperlinks
\usepackage{url}            % simple URL typesetting
\usepackage{booktabs}       % professional-quality tables
\usepackage{amsfonts}       % blackboard math symbols
\usepackage{nicefrac}       % compact symbols for 1/2, etc.
\usepackage{microtype}      % microtypography
\usepackage{xcolor}         % colors
\usepackage{graphicx}
\usepackage{multirow}
\usepackage{braket}
\usepackage{algorithm}
\usepackage{algorithmic}
\usepackage{bm}
\usepackage{hhline}
\usepackage[accsupp]{axessibility}  % Improves PDF readability for those with disabilities.

\begin{document}
% \linenumbers
\pagestyle{headings}
\mainmatter
\def\ECCVSubNumber{2717}  % Insert your submission number here

\title{Distilling Object Detectors with Global Knowledge} % Replace with your title

\begin{comment}
\titlerunning{2717 \ECCVSubNumber} 
\authorrunning{2717 \ECCVSubNumber} 
\author{Anonymous ECCV submission}
\institute{Paper ID \ECCVSubNumber}
\end{comment}
%******************

%\begin{comment}
\titlerunning{Distilling Object Detectors with Global Knowledge}
\author{Sanli Tang\inst{1}\textsuperscript{$\star$} \and Zhongyu Zhang\inst{1}\thanks{Authors contributed equally. $\dagger$ Corresponding authors.} \and Zhanzhan Cheng\inst{1}\textsuperscript{$\dagger$} \and Jing Lu\inst{1} \and Yunlu Xu\inst{1} \and Yi Niu\inst{1} \and Fan He\inst{2}}
\authorrunning{Tang et al.}
\institute{Hikvision Research Institute, Hanzhou, China \and Institute of Image Processing and Pattern Recognition, Shanghai Jiao Tong University, Shanghai, China \\ 
\email{\{tangsanli,zhangzhongyu,chengzhanzhan,lujing6,xuyunlu,niuyi\}@hikvision.com \\ hf-inspire@sjtu.edu.cn}
}
%\end{comment}
%******************

\maketitle

\begin{abstract}
Knowledge distillation learns a lightweight student model that mimics a cumbersome teacher. Existing methods regard the knowledge as the feature of each instance or their relations, which is the instance-level knowledge only from the teacher model, i.e., the \emph{local} knowledge. However, the empirical studies show that the \emph{local} knowledge is much noisy in object detection tasks, especially on the blurred, occluded, or small instances. Thus, a more intrinsic approach is to measure the representations of instances w.r.t. a group of \emph{common} basis vectors in the two feature spaces of the teacher and the student detectors, i.e., \emph{global} knowledge. Then, the distilling algorithm can be applied as space alignment. To this end, a novel prototype generation module (PGM) is proposed to find the \emph{common} basis vectors, dubbed \emph{prototypes}, in the two feature spaces. Then, a robust distilling module (RDM) is applied to construct the global knowledge based on the prototypes and filtrate noisy local knowledge by measuring the discrepancy of the representations in two feature spaces. Experiments with Faster-RCNN and RetinaNet on PASCAL and COCO datasets show that our method achieves the best performance for distilling object detectors with various backbones, which even surpasses the performance of the teacher model. We also show that the existing methods can be easily combined with global knowledge and obtain further improvement. Code is available: \url{https://github.com/hikvision-research/DAVAR-Lab-ML}.
\keywords{Object Detection, Knowledge Distillation}
\end{abstract}

\section{Introduction}
Object detectors can be enhanced by applying larger networks \cite{ResNet,Deeper}, which, however, will increase the storage and computational cost. A promising solution for finding the sweet spot between efficiency and performance is knowledge distillation (KD) \cite{Compression,Hinton}, which learns a lightweight student that mimics the behaviors of a cumbersome teacher.

\begin{figure}[t]
\centering 
\begin{minipage}[t]{0.54\textwidth}
\centering
\includegraphics[width=6.2cm]{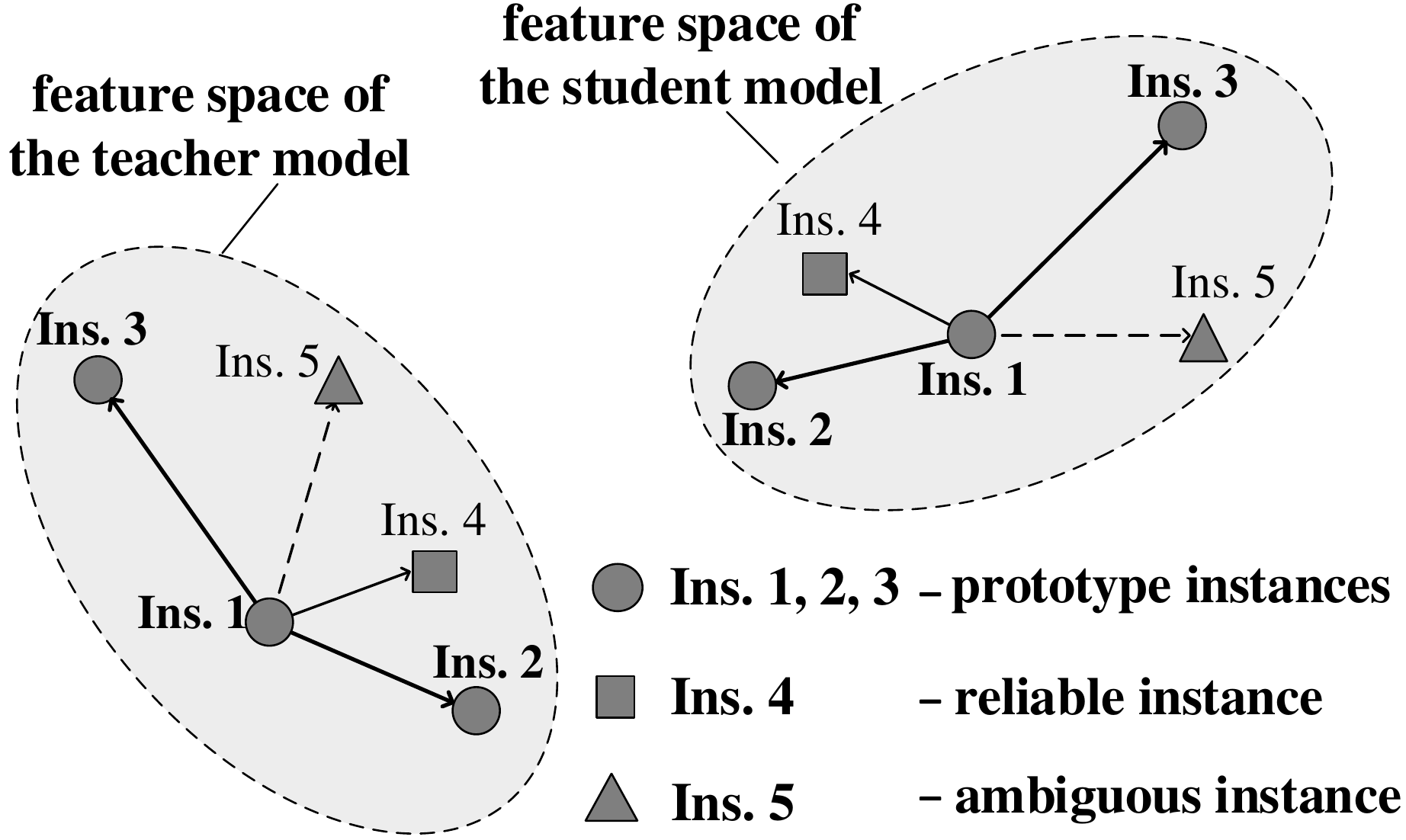}
\end{minipage}
\begin{minipage}[t]{0.44\textwidth}
\centering
\includegraphics[width=5.2cm]{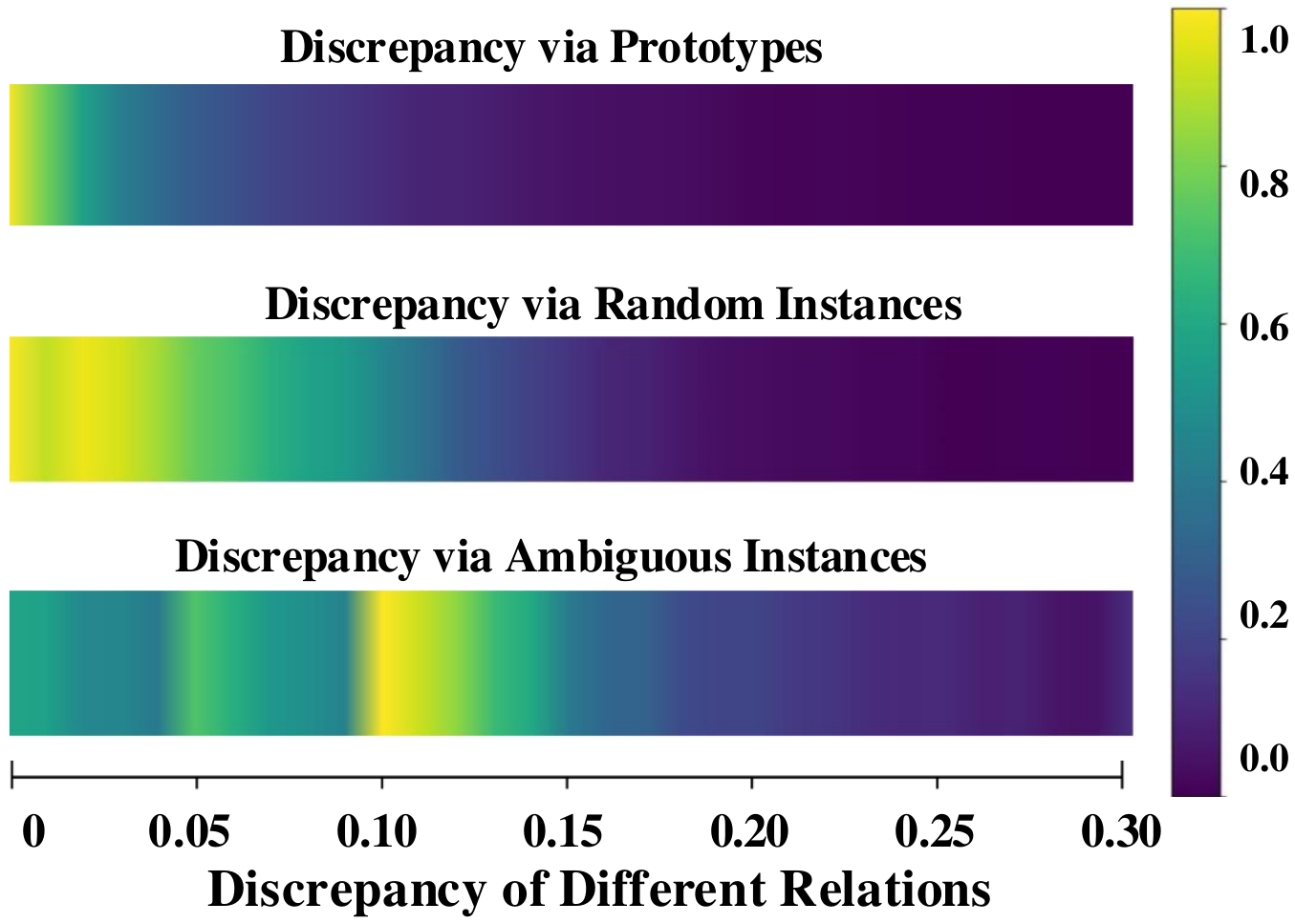}
\end{minipage}
\caption{\textbf{Left}: the prototypes are representative and play roles as a common group of basis vectors in \emph{TS-space}. Although the absolute location of \emph{Ins. 4} is different in \emph{TS-space}, its representations, e.g., the relations, w.r.t. prototypes are similar while \emph{Ins. 5} shows the representation of much dissimilar. \textbf{Right}: on COCO dataset \cite{COCO} with Faster-RCNN detector \cite{Faster}, we show the discrepancy of relations between instances and three types of basis in \emph{TS-space}. 10 instances are selected for each class as the bases and others are used for measuring the discrepancy of relations in \emph{TS-space}. The relations with \emph{prototypes} show much smaller discrepancy than others.}
\label{fig-example}
\end{figure}

The knowledge can be known to be formed in three categories \cite{Survey}: feature-based knowledge \cite{Fitnets,Gift,Boundaries,FGFI,FBKD}, response-based knowledge \cite{Hinton,Mimicking,RKD}, and relation-based knowledge \cite{RKD,RKDGraph,Similarity,Correlation,GIBox}. Such knowledge can be treated as the \emph{local} knowledge, since only the instance-level knowledge from a single feature space, e.g., the teacher's, is considered.
Based on these knowledge, existing methods design their distilling algorithms for object detection tasks based on some prior senses, e.g., the foreground regions \cite{FGFI}, the decoupled background regions \cite{DecoupledFeature}, the attention guided regions \cite{FBKD,ICD}, or the discrepancy regions \cite{GIBox,PFI}. However, we find that the local knowledge is of much discrepancy between the teacher and the student in object detection tasks, especially on the ambiguous instances which are blur, truncated, or small. This is because features of ambiguous instances are susceptible to the small disturbance in feature spaces of the teacher and the student.
Thus, the distilling process will suffer from the noisy local knowledge, e.g., the false positives and the localization errors, and lead to sub-optimal.

The main concerns on relieving the effect of noisy local knowledge are two folds: constructing reliable global knowledge and applying robust distilling algorithms.
By viewing knowledge as the representation of feature space, a more intrinsic approach is to find a group of common basis vectors in both the feature spaces of the teacher and the student detectors. In this way, the \emph{global} knowledge can be formed by representing the instances w.r.t. these basis vectors. Then, a more robust distilling algorithm can be designed by measuring the discrepancy of the representations in the two feature spaces. 
Hereafter, we name the two feature spaces of the teacher and the student detector as \emph{TS-space} and the common basis vectors of the TS-space as \emph{prototypes}.

In Fig. \ref{fig-example} (left), we illustrate that: (1) the representations of normal instances w.r.t. the prototypes are of the little discrepancy between two feature spaces, e.g., the \emph{Ins.4}; (2) the discrepancy of the ambiguous instances is much larger than others, e.g., the \emph{Ins. 5}.
In Fig. \ref{fig-example} (right), we show the statistic analysis of the discrepancy of the instance representations in \emph{TS-space} on the COCO dataset.
Notice that each instance is represented by a pair of features in the \emph{TS-space}. Thus, we first measure the cosine similarity between the bases and each of the other instances in the \emph{TS-space}, and then calculate the discrepancy by $l_1$ distance as shown by the abscissa. In Fig. \ref{fig-example} (right), the discrepancy of relations between prototypes and other instances is much smaller than other bases, which shows a more promising representation of the knowledge in \emph{TS-space}.

Based on the above considerations, we first propose a prototype generation module (PGM) to find a group of common basis vectors as the prototypes in \emph{TS-space}. It selects the prototypes according to minimizing the reconstruction errors of the instances in the two feature spaces, which is inspired by the dictionary learning \cite{Dictionary,Dictionary2,MP}.
Then, a robust distillation module (RDM) is designed for robust knowledge construction and transfer. Based on the prototypes, the global knowledge is formed by representing the instances under the prototypes, which shows a smaller gap between the two spaces as in Fig. \ref{fig-example} (right). The discrepancy of the representations in \emph{TS-space} can also be regarded as an ensemble of the two models to mitigate noisy local knowledge transferring when distilling. 
Experiments are carried out with both single-stage (RetinaNet \cite{RetinaNet}) and two-stage detectors (Faster R-CNN \cite{Faster}) on Pascal VOC \cite{VOC} and COCO \cite{COCO} benchmarks. Extensive experimental results show that the proposed method can effectively improve the performance of knowledge distillation, which achieves new remarkable performance. We also show the existing methods can be further improved by the prototypes with global and local knowledge.

\section{Related Works}
\subsection{Object Detection}
Existing object detection methods based on deep neural networks can be divided into anchor-based and anchor-free detectors. 
The anchor-based detectors use the preset boxes as anchors, which are trained to classify their categories and regress the offsets of coordinates.
They can be further divided into multi-stage \cite{Fast,Faster} and single-stage \cite{Yolo,SSD,DSSD} detectors. 
As the representative multi-stage detector, Faster R-CNN \cite{Faster} uses a region proposal network to generate proposals that probably contain objects and then predicts their categories and refines the proposals in the second stage. 
Considering the large computation cost of the multi-stage detectors, YOLO \cite{Yolo}, as the representative single-stage detector, is proposed to use a fully convolutional network to predict both the bounding boxes and categories. It is further improved by applying feature pyramid \cite{SSD}, deconvolutional layers \cite{DSSD} and focal loss \cite{RetinaNet} to treat the various object scales, the semantic information of features, and the unbalance of positives and negatives, respectively. Many anchor-free detectors \cite{FCOS,Selective} are proposed to avoid empirically setting and tedious calculation of the anchors. Although applying deeper and wider networks can often improve the performance of detectors, it is too computationally expensive in many resource-limited applications.
\subsection{Knowledge Distillation}
Knowledge distillation \cite{Hinton,PAAD,COFD} is proposed by Hinton et al. \cite{Hinton} in the image classification task to transfer knowledge of a cumbersome teacher model into a compact student model. 
There are two main aspects of knowledge distillation: knowledge construction and knowledge transfer.
For the first aspect, knowledge mainly consists of three types \cite{Survey}: the feature-based knowledge, i.e., activations of intermediate feature \cite{Fitnets,Neuron,Gift,Boundaries}, the relation-based knowledge, i.e., structures in the embedding space \cite{RKD,Graph,Contrastive,Similarity,Correlation}, and the response-based knowledge, i.e., the soft target of the output layers \cite{Hinton}. 
For the second aspect to effectively transfer the knowledge to the student. \cite{Hinton} applies a temperature factor to control the softness of the probability distribution over classes. \cite{Fitnets} adds a regression layer as a bridge to match dimensions of the features. Such knowledge can be viewed as local knowledge since only the instance-level knowledge in the single feature space, e.g., the teacher's is considered.

For distilling an object detector \cite{Efficient,DecoupledFeature,Localization,FBKD,PFI,ICD}, more attention is paid on \emph{constructing knowledge} due to the extreme imbalance over the foreground/background areas and the numbers of instances among different classes. 
\cite{FGFI} aims at keeping the balance between foreground and background features by distilling on the areas around ground-truth boxes, while \cite{Mimicking} distills on high-level features within the equivalently sampled foreground and background proposals by referring to the ground-truth boxes. \cite{GIBox} is recently proposed to distill features in anchors where there are the most discrepancies of confidence between the student and the teacher model. \cite{Adaptive} proposes to gradually reduce the distillation penalty to balance the two targets of detection and distillation. 
However, existing methods regard the activations or the relations between all instances as the local knowledge to distill object detectors, which suffers from the noises, e.g., the ambiguous instances or the detection errors from the teacher.

\section{Method}
\begin{figure}[tbp]
\centering
\includegraphics[width=12.2cm]{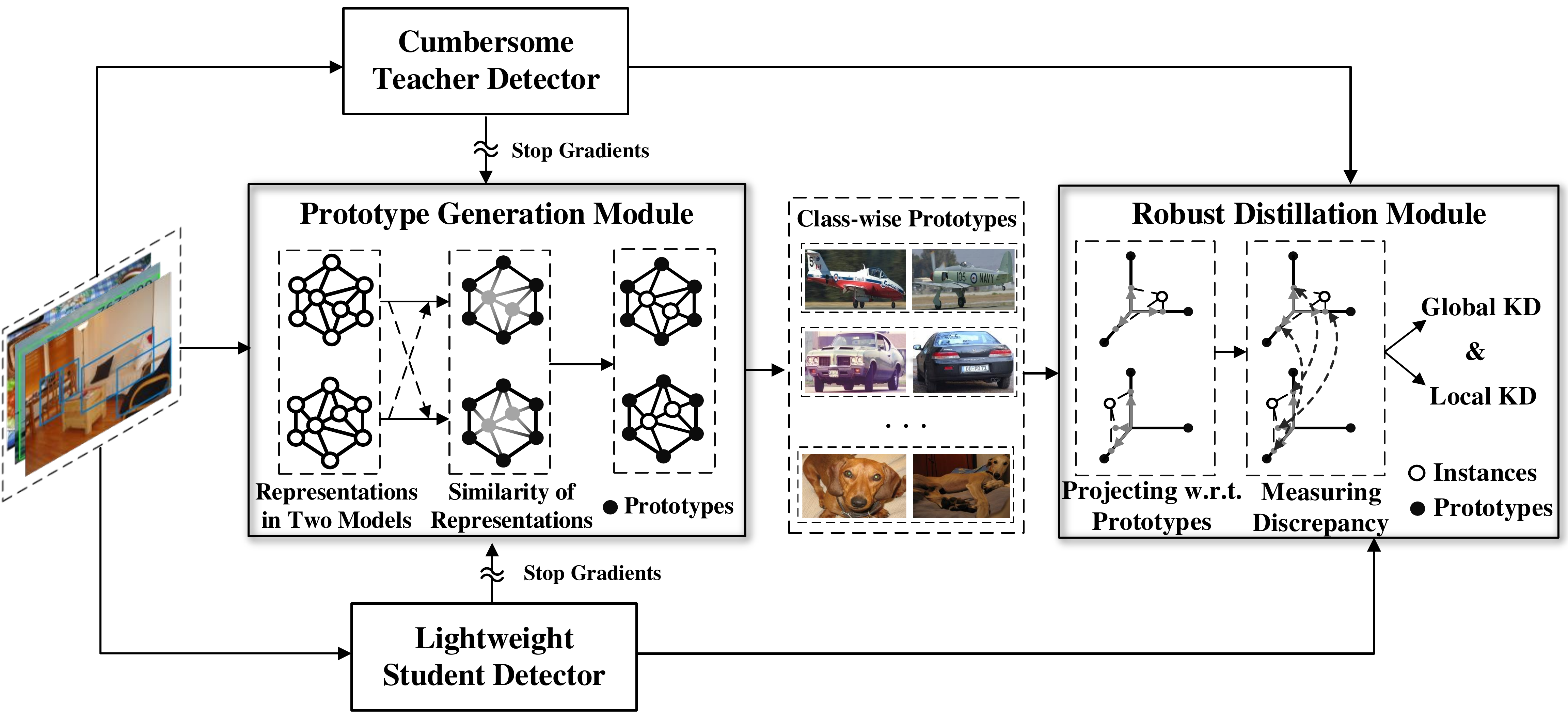}
\caption{The proposed framework for distilling object detectors with global knowledge. A prototype generation module (PGM) is first deployed to find the prototypes for each class based on the similarity of their representations in \emph{TS-space}. A robust distillation module (RDM) is then designed to construct reliable global knowledge w.r.t. the prototypes and measure their discrepancy for robust knowledge distillation.}
\label{fig-framework}
\end{figure}

In this section, we detail the proposed framework for distilling object detectors with global knowledge. As shown in Fig. \ref{fig-framework}, the overall framework consists of two modules: a prototype generation module (PGM) to find class-wise prototypes for bridging the two feature spaces, and a robust distilling module (RDM) to construct and distill the reliable global knowledge based on the prototypes.
\subsection{Prototype Generation Module} 
The knowledge of a deep model can be viewed as the representation of its feature space, which can be approximated by a small set of basis vectors from the view of dictionary learning \cite{Dictionary,Dictionary2}, as shown in Fig \ref{fig-example} (left).
Concretely, let ${\bm F}=\{{\bm f_i}\}_{i=1}^N\in\mathbb{R}^{D{\times}N}$ be features of $N$ instances in the feature space of $D$ dimensions. The $K$ $(K{\ll}N)$ basis vectors $\bm G=\{{\bm g_i}\}_{i=1}^K{\subset}{\bm F}$ of a single feature space can be selected by minimizing the reconstruction errors of all instances:
\begin{equation}
    {\bm G},\ {\bm W} = \mathop{\arg\min}_{{\bm G},{\bm W}}||{\bm F} - {\bm G}{\bm W}||_2^2 + \lambda||{\bm W}||_1^2,
\end{equation}
where $\bm W\in\mathbb{R}^{K{\times}N}$ is the representation of all samples $\bm F$ w.r.t. the basis vectors ${\bm G}$. The last regularized term weighted by $\lambda$ helps to learn a sparse $\bm W$, which makes the basis vectors ${\bm G}$ representative.

In the knowledge distillation task, there are two different feature spaces that are represented by the teacher and student detectors, namely \emph{TS-space}. Thus, a more intrinsic approach is to find a group of common basis vectors in \emph{TS-space}, which bridge the gap between the two spaces and reduce the difficulty of distillation. Following the above considerations, the prototype generation module (PGM) aims at finding $K$ instances as the basis vectors, dubbed \emph{prototypes}. In this way, other instances can be represented by prototypes with minimum reconstruction errors in each of the feature spaces. Meanwhile, the representing discrepancy based on the prototypes between two feature spaces should also be small such that it is easier to transform one feature space to another.

Let ${\bm F_t}=\{\bm{f}_i^t\}_{i=1}^N\in\mathbb{R}^{D_t \times N}$ and ${\bm F_s}=\{\bm{f}_i^s\}_{i=1}^N\in\mathbb{R}^{D_s \times N}$ be the $N$ instances in the feature spaces of the teacher and the student detectors, respectively. $D_t$ and $D_s$ are the dimensions of the two feature spaces. 
The $K$ prototypes can be grouped as
$({\bm G_t}, {\bm G_s})=\{(\bm{g}_i^t, \bm{g}_i^s)\}_{i=1}^K$, where ${\bm G_t}={\bm F}_{{\bm i}(t)}\subset{\bm F_t}$ and ${\bm G_s}={\bm F}_{{\bm i}(s)}\subset{\bm F_s}$ are the subset of all instances in \emph{TS-space}.
${\bm i}(s)$, ${\bm i}(t)$ are the indexing sets.

Notice that the prototypes are the common basis vectors of \emph{TS-space}.
Thus, they can be generated from all the instances by minimizing the reconstruction errors in both of the two feature spaces as well as a regularization of the representing consistency with a trade-off weight $\lambda$:
\begin{equation}
    \begin{split}
     & ||{\bm F_t}-{\bm F}_{{\bm i}(t)}{\bm W_t}||_2^2+||{\bm F_s}-{\bm F}_{{\bm i}(s)}{\bm W_s}||_2^2+\lambda||\bm W_s-\bm W_t||_2^2 \\
     &\quad \quad \quad  {\rm s.t.} \quad   {\bm i}(s)={\bm i}(t) \quad {\rm and} \quad  |{\bm i}(s)|=|{\bm i}(t)|=K
    \end{split}
\label{eq-prototype-objective}
\end{equation}
$\bm W_t=\{w_{j,i}^t\}_{K{\times}N}$ and $\bm W_s=\{w_{j,i}^s\}_{K{\times}N}$ are the representations, i.e., coordinates, of the $N$ instances w.r.t. the $K$ prototypes in the feature spaces of the teacher and the student. The last term in Eq. \ref{eq-prototype-objective} requires the representations of instances in \emph{TS-space} are similar such that the discrepancy of the relations is small, as illustrated in Fig. \ref{fig-example} (right). The constraint ${\bm i}(s)={\bm i}(t)$ requires that a prototype is indeed one instance represented in two feature spaces, and total $K$ prototypes are selected. In this way, representations of instances w.r.t. the prototypes can be regarded as approximations of the feature space, i.e., the global knowledge, as shown in Fig. \ref{fig-example} (left), which allows penalizing the difference of the relations between instances and prototypes in two feature spaces for knowledge transfer. Besides, the discrepancy of relations w.r.t. the prototypes can be used as the robustness cue for knowledge transfer, as illustrated in Fig. \ref{fig-example}.

We show an approximate solution of the problem in Eq. \ref{eq-prototype-objective} through a variant of matching pursuit \cite{MP}, which is indeed a greedy algorithm yet very efficient. To select the $(n+1)^{\rm th}$ prototype $(\bm{g}_{n+1}^t,\bm{g}_{n+1}^s)$, we first define the residuals ${\bm r}_{n,i}^t$ and ${\bm r}_{n,i}^s$ w.r.t. the selected $n$ prototypes as follows:
\begin{equation}
{\bm r}_{n,i}^t{\triangleq}{\bm f}_i^t-\sum_{k=1}^{n}{\bm g}_k^t w_{k,i}^t,\quad {\bm r}_{n,i}^s{\triangleq}{\bm f}_i^s-\sum_{k=1}^{n}{\bm g}_k^s w_{k,i}^s.
\label{eq-r}
\end{equation}
The objective in Eq. \ref{eq-prototype-objective} w.r.t. the $(n+1)^{\rm th}$ prototype can be written by 
\begin{equation}
\mathcal{L}_{n+1}=\sum_{i=1}^N||{\bm r}_{n+1,i}^t||_2^2+\sum_{i=1}^N||{\bm r}_{n+1,i}^s||_2^2 +\lambda\sum_{i=1}^N\sum_{k=1}^{n+1}(w_{k,i}^t-w_{k,i}^s)^2.
\label{eq-l}
\end{equation}
The optimal $w_{n+1,i}^t$ and $w_{n+1,i}^s$ can be obtained by making the derivative of the $\mathcal{L}_{n+1}$ with respect of $w_{n+1,i}^t$ and $w_{n+1,i}^s$ to zero. Then, we have
\begin{equation}
w_{n+1,i}^t=\frac{\Braket{{\bm r}_{n,i}^t,{\bm g}_{n+1}^t}+{\lambda}w_{n+1,i}^s}{\lambda+||{\bm g}_{n+1}^t||_2^2},\quad
w_{n+1,i}^s=\frac{\Braket{{\bm r}_{n,i}^s,{\bm g}_{n+1}^s}+{\lambda}w_{n+1,i}^t}{\lambda+||{\bm g}_{n+1}^s||_2^2}.
\label{eq-w}
\end{equation}
We detail the derivation and show the closed-form solution of Eq. \ref{eq-w} in the supplemental materials, where we also show more analysis about the relationship between global knowledge and relation-based knowledge. The overall algorithm for generating prototypes is summarized in Alg. \ref{alg-pgm}. Notice that we separately generate prototypes for each class.

\begin{algorithm}[tb] 
\caption{Algorithm for selecting prototypes in PGM.} 
\label{alg-pgm} 
\textbf{Input}: \\
$\{({\bm f}^t_i, {\bm f}^s_i)\}_{i=1}^N$: features of $N$ instances in \emph{TS-space};  \\
\textbf{Parameter}: \\
$K$: number of prototypes to be selected; \\
$\lambda$: regularization weight; \\
\textbf{Output}: \\
$\mathcal{I}$: index set of the prototypes   

\begin{algorithmic}[1]
\STATE initialize $n=0$, the residuals ${\bm r}_{0,i}^t={\bm f}_i^t$, and ${\bm r}_{0,i}^s={\bm f}_i^s\quad\forall i=1,\cdots,N$;
\WHILE{$n < K$}
\STATE compute the optimal $w_{n+1,i}^s$ and $w_{n+1,i}^t$ by Eq. \ref{eq-w};
\STATE compute the $\mathcal{L}_{n+1}^k$ with Eq. \ref{eq-l} for each instance by setting ${\bm g}_{n+1}^s={\bm f}_k^s$ and ${\bm g}_{n+1}^t={\bm f}_k^t$, $\forall k=1,\cdots,N$ ;
\STATE append the index $k^*$ into $\mathcal{I}$ where $k^*=\arg\min_k\left\{\mathcal{L}_{n+1}^k\right\}\ \forall\ ({\bm g}_k^s,{\bm g}_k^t)\in\{({\bm f}_i^s,{\bm f}_i^t)\}_{i=1}^N$; set $\bm{g}_{n+1}^t=\bm{f}_{k^*}^t$ and $\bm{g}_{n+1}^s=\bm{f}_{k^*}^s$;
\STATE update the residuals ${\bm r}_{n+1,i}^t$ and ${\bm r}_{n+1,i}^s$ by Eq. \ref{eq-r};
\STATE set $n=n+1$;
\ENDWHILE
\STATE \textbf{Return}: $\mathcal{I}$
\end{algorithmic} 
\end{algorithm}

\subsection{Robust Distillation Module} 
In this section, we focus on \emph{global knowledge construction} and \emph{robust knowledge transferring} by a robust distillation module (RDM) based on the prototypes.

\textbf{Identifying the knowledge}. 
By referring to the prototypes, the global knowledge, i.e., the representations of instances on the common basis vectors in the two feature spaces, can be naturally constructed by measuring the representation between the instances and the prototypes. 

Specifically, let the features of the $j^{\rm th}$ instance in the $i^{\rm th}$ image be ${\bm f}_{i,j}^t$ and ${\bm f}_{i,j}^s$ in the feature spaces of the teacher and the student detectors, respectively. For an instance with a pair of features $({\bm f}_{i,j}^t,{\bm f}_{i,j}^s)$ in \emph{TS-space}, they can be separately projected onto the common basis vectors $(\bm{G}_t,\bm{G}_s)$ in each space as $\bm{\Lambda}_{i,j}^t=\mathcal{P}_{\bm{G}_t}(\bm{f}_{i,j}^t)$ and $\bm{\Lambda}_{i,j}^s=\mathcal{P}_{\bm{G}_s}(\bm{f}_{i,j}^s)$. $\mathcal{P}$ is the projection function. The project coefficients $\bm{\Lambda}_{i,j}^t$ and $\bm{\Lambda}_{i,j}^s$ can be calculated exactly the same as in Eq. \ref{eq-w}.

Thus, the global knowledge can be transferred by minimizing:
\begin{equation}
\mathcal{L}_{\rm global}=\frac{1}{2NK}\sum_{i=1}^n\sum_{j=1}^{n_i}\sigma_{i,j}||\bm{\Lambda}_{i,j}^s-\bm{\Lambda}_{i,j}^t||_2^2,
\label{eq-pair}
\end{equation}
where $N=\sum_{i=1}^nn_i$ are the total number of instances. $n$ is the number of images and $n_i$ is the number of instances in the $i$-th image. $\sigma_{i,j}$ is the weight that reveals how reliable the knowledge is and will be discussed later. 

For the local feature-based knowledge, we follow \cite{FGFI} identifying the knowledge as the features of the regions that overlap with any ground-truth boxes larger than a threshold. Thus, the local feature-based knowledge can be defined as:
\begin{equation}
\mathcal{L}_{\rm local}^{\rm feat}=\frac{1}{2N}\sum_{i=1}^n\sum_{j=1}^{n_i}\sigma_{i,j}||\mathcal{H}(\bm{f}_{i,j}^s)-\bm{f}_{i,j}^t||_2^2, 
\label{eq-point}
\end{equation}
where $\mathcal{H}$ is an adaptation function, e.g., a $1 \times 1$ convolutional layer with ReLU activation in our paper, that transforms the features of the student into the feature space of the same dimensions as the teacher's. 

For the local response-based knowledge, we use the proposals and apply the RoI-align \cite{Faster} to get the prediction inside the regions. The KL-divergence weighted by $\sigma_{i,j}$ is used on the predicting logits between the teacher and the student, and denoted as $\mathcal{L}_{\rm local}^{\rm resp}$.

\textbf{Robustly distilling the knowledge}. Since the knowledge from the teacher might be noisy, especially on ambiguous instances, a robust knowledge transferring approach is required to distinguish noisy knowledge and mitigate transferring them to the student. Inspired from co-teaching \cite{MentorNet,Decoupling,Co-teaching} to alleviate the noise from multiple views, the student might also have a voice in discriminating the noisy knowledge. Based on the observations that reliable knowledge should have similar representations under the measurement from two models, shown in Fig. \ref{fig-example}, the robustness of knowledge can be estimated by the discrepancy of representations in \emph{TS-space}. Thus, the weight $\sigma_{i,j}$ for fine-grained knowledge distillation can be approximated as
\begin{equation}
\label{eq-weight}
\sigma_{i,j}=1-||\bm{\Lambda}_{i,j}^s-\bm{\Lambda}_{i,j}^t||_2.
\end{equation}
$\sigma_{i,j}$ describes the similarity of the representations between the instance and the prototypes in the \emph{TS-space}.
It is indeed heavily related to the last term in Eq. \ref{eq-prototype-objective}, where we concentrate more on the instances with small discrepancy w.r.t. the prototypes for both global and local knowledge transfer.

\begin{algorithm}[tb]
\caption{The proposed knowledge distilling process.}
\label{alg-r2i}
\textbf{Input}: teacher detector $\mathcal{T}$, student detector $\mathcal{S}$, prototype updating period $T$ and maximum training epochs $T_m$.\\
\begin{algorithmic}[1] %[1] enables line numbers
\STATE let $e$ be the current training epoch and set $e=0$;
\WHILE{$e<T_m$}
\IF {$\rm mod(e, T)==0$}
\STATE extract features of instances ${\bm F_t}$ and ${\bm F_s}$ from the teacher $\mathcal{T}$ and current student (at $e$-th epoch) $\mathcal{S}^e$, respectively;
\STATE updating and bootstrapping prototypes $(\bm{G}_t, \bm{G}_s)$ for each class by minimizing Eq. \ref{eq-prototype-objective} based on $\mathcal{T}$ and $\mathcal{S}^e$ (see Alg. \ref{alg-pgm});
\ENDIF
\STATE training the student detector for one epoch by minimizing Eq. \ref{eq-objective};
\STATE set $e=e+1$;
\ENDWHILE
\end{algorithmic}
\end{algorithm}

\subsection{Optimization}
The overall objective for distilling object detectors can be summarized as:
\begin{equation}
\mathcal{L}_{\rm kd}=\mathcal{L}_{\rm det}+\alpha_1\mathcal{L}_{\rm global}+\alpha_2\mathcal{L}_{\rm local}^{\rm feat}+\alpha_3\mathcal{L}_{\rm local}^{\rm resp},
\label{eq-objective}
\end{equation}
where $\mathcal{L}_{\rm det}$ is the original detection objective defined by the student detector. $\alpha_1$, $\alpha_2$, and $\alpha_3$ weigh the global and local knowledge transfer. For detectors with FPN \cite{FPN} using multiple feature maps for prediction, we independently apply the PGM and the RDM on each of the feature maps. Since the student is gradually optimized and the relations are changed, the prototypes should be updated when training the student. For efficiency, the prototypes are bootstrapped and updated every $T$ epochs. Both the student and the teacher detectors are pre-trained on the task-relevant dataset to extract features of instances and generate the prototypes. The overall proposed distilling algorithm is summarized in Alg. \ref{alg-r2i}

\begin{table}[tbp]
\caption{Knowledge distillation results on COCO dataset with different detectors. Some results are missing since we cannot find the performance report in their papers.}
\label{tb-coco}
\centering
\begin{tabular}{c|c|cc|ccc|c|ccc}
\toprule
\textbf{Method}&\textbf{mAP}&\textbf{AP}$_{50}$&\textbf{AP}$_{75}$&\textbf{AP}$_s$&\textbf{AP}$_{m}$&\textbf{AP}$_{l}$&\textbf{mAR}&\textbf{AR}$_s$&\textbf{AR}$_m$&\textbf{AR}$_l$\\
\midrule
Faster-Res101 (teacher) & 39.8 & 60.1 & 43.3 & 22.5 & 43.6 & 52.8 & 53.0 & 32.8 & 56.9 & 68.6 \\
Faster-Res50 (student)& 38.4 & 59.0 & 42.0 & 21.5 & 42.1 & 50.3 & 52.0 & 32.6 & 55.8 & 66.1 \\
FGFI \cite{FGFI} & 39.3 & 59.8 & 42.9 & 22.5 & 42.3 & 52.2 & 52.4 & 32.2 & 55.7 & 67.9 \\
DeFeat \cite{DecoupledFeature} & 40.3 & 60.9 & \textbf{44.0} & 23.1 & 44.1 & \textbf{53.4} & 53.7 & 33.3 & 57.7 & 69.1 \\
FBKD \cite{FBKD} & 40.2 & 60.4 & 43.6 & 22.8 & 43.8 & 53.2 & 53.4 & 32.7 & 57.1 & 68.8 \\
GID \cite{GIBox} & 40.2 & 60.8 & 43.6 & \textbf{23.6} & 43.9 & 53.0 & 53.7 & 33.6 & 57.7 & 68.6 \\
Ours & \textbf{40.6} & \textbf{61.0} & \textbf{44.0} & 23.4 & \textbf{44.4} & 53.3 & \textbf{53.8} & \textbf{33.9} & \textbf{57.9} & \textbf{69.2} \\
\midrule
Retina-Res101 (teacher) & 38.9 & 58.0 & 41.5 & 21.0 & 42.8 & 52.4 & 54.8 & 33.4 & 59.3 & 71.2 \\
Retina-Res50 (student) & 37.4 & 56.7 & 39.6 & 20.0 & 40.7 & 49.7 & 53.9 & 33.1 & 57.7 & 70.2 \\
FGFI \cite{FGFI} & 38.6 & 58.7 & 41.3 & 21.4 & 42.5 & 51.5 & 54.6 & 34.7 & 58.2 & 70.4 \\
GID \cite{GIBox} & 39.1 & \textbf{59.0} & 42.3 & \textbf{22.8} & 43.1 & 52.3 & 55.3 & \textbf{36.7} & 59.1 & 71.1 \\
DeFeat \cite{DecoupledFeature} & 39.3 & 58.2 & 42.1 & 21.7 & 42.9 & 52.9 & 55.1 & 33.9 & 59.6 & 71.5 \\
FBKD \cite{FBKD} & 39.3 & 58.8 & 42.0 & 21.2 & 43.2 & 53.0 & 55.4 & 34.6 & 59.7 & 72.2 \\
FR \cite{FR} & 39.3 & 58.8 & 42.0 & 21.5 & 43.3 & 52.6 &  - & - & - & - \\
PFI \cite{PFI} & 39.6 & - & - & 21.4 & 44.0 & 52.5 & - & - & - & - \\
Ours & \textbf{39.8} & 58.6 & \textbf{42.6} & 21.8 & \textbf{43.5} & \textbf{53.5} & \textbf{55.8} & 34.1 & \textbf{60.0} & \textbf{72.2} \\
\bottomrule
\end{tabular}
\end{table}

\section{Experiments}
\label{sec-exp}
We perform experiments with the representative single-stage and two-stage detectors, namely, RetinaNet \cite{RetinaNet} and Faster R-CNN \cite{Faster} on the PASCAL VOC \cite{VOC} and COCO \cite{COCO} detection benchmarks. We follow the common settings that use both VOC 07 and 12 \emph{trainval} split for training and VOC 07 \emph{test} split for test. For the COCO dataset, the \emph{train} split are used for training while the \emph{val} split are used for test. Unless otherwise specified, the hyper-parameters are set as $K=10$, $\lambda=10$ and $T=1$. The distilling weights $\alpha_1$, $\alpha_2$, and $\alpha_3$ are set to 1.0, 1.0, 5.0, respectively. The student detector is trained through 2$\times$ learning schedule on 8 Tesla V100 32G GPUs. The input images are resized as large as $1333\times800$ while keeping the aspect ratio. Other standard augmentations, e.g., the photometric distortion, are applied as the settings in MMDetection \cite{mmdet}.
The ResNet101 and ResNet50 \cite{ResNet} backbones are used for the teacher and the student detectors, respectively. We also validate our methods with larger teachers, e.g., Cascade Mask R-CNN \cite{CascadeRCNN} with ResNext-101 \cite{ResNext}. More implementation details are included in the supplemental material.

\subsection{Comparison with existing methods on VOC and COCO datasets}

We first evaluate our method on VOC and COCO datasets with the representative two-stage detector (Faster R-CNN) and single-stage detector (RetinaNet). As shown in Table \ref{tb-coco} and Table \ref{tb-voc}, all student models are significantly improved by our knowledge distillation algorithm, e.g., 2.2\% and 2.4\% mAP on the COCO dataset and 2.5\% and 2.5\% mAP on the VOC dataset for both detectors. Moreover, they even surpass the teacher detector within a large margin, e.g., 0.8\%, 0.9\% on COCO dataset for both detectors. As we form the global knowledge as the ensemble of both the student and the teacher detectors and use common basis vectors to bridge the two feature spaces for distilling, the proposed method shows more potential to achieve a further gain compared to the teacher detectors.

We also compare our method with the SOTA detection distillation methods with the same teacher and student detectors. Table \ref{tb-coco} and Table \ref{tb-voc} show that the proposed method achieves best mAP on COCO and VOC datasets. Notice that GID \cite{GIBox} applies all the three types of local knowledge, i.e., feature-based, relation-based, and response-based knowledge for distilling. However, the proposed method shows further improvement on both COCO and VOC datasets, e.g, 0.7\% mAP gain for distilling the RetinaNet. It reveals that distilling the knowledge by forcing the student to absolutely behave the same as the teacher still leads to sub-optimal since the local knowledge represented by the ambiguous instances is hard to transfer and will hurt the distillation. The proposed method shows a more promising way by looking for a group of common basis vectors, i.e., the prototypes, for bridging the gap of the two feature spaces and forming as well as distilling the global knowledge based on the prototypes in a more robust way. Moreover, the results in Table \ref{tb-coco} and Table \ref{tb-voc} demonstrate that our method is capable to be applied to various detection frameworks.

\begin{table}[t]
\caption{Knowledge distillation results on Pascal VOC dataset with different detectors.}
\label{tb-voc}
\centering
\setlength{\tabcolsep}{3mm}
\begin{tabular}{c|ccc|ccc}
\toprule
\multirow{2}{*}{\textbf{Method}} & \multicolumn{3}{c|}{\textbf{Faster R-CNN Res101-50}}  & \multicolumn{3}{c}{\textbf{RetinaNet Res101-50}} \\
\cline{2-7} & \textbf{mAP} & \textbf{AP}$_{50}$ & \textbf{AP}$_{75}$ & \textbf{mAP} & \textbf{AP}$_{50}$ & \textbf{AP}$_{75}$\\
\midrule
teacher & 56.3 & 82.7 & 62.6 & 58.2 & 82.0 & 63.0 \\
student & 54.2 & 82.1 & 59.9 & 56.1 & 80.9 & 60.7 \\
FitNet \cite{Fitnets} & 55.0 & 82.2 & 61.2 & 56.4 & 81.7 & 61.7 \\
FGFI \cite{FGFI} & 55.3 & 82.1 & 61.1 & 55.6 & 81.4 & 60.5 \\
FBKD \cite{FBKD} & 55.4 & 82.0 & 61.3 & 56.7 & 81.9 & 61.9 \\
ICD \cite{ICD} & 56.4 & 82.4 & 63.4 & 57.7 & \textbf{82.4} & 63.5 \\
GID \cite{GIBox} & 56.5 & 82.6 & 61.6 & 57.9 & 82.0 & 63.2 \\
Ours & \textbf{56.7} & \textbf{82.9} & \textbf{61.9} & \textbf{58.6} & \textbf{82.4} & \textbf{64.2} \\
\bottomrule
\end{tabular}
\end{table}

\subsection{Effects of the prototypes in robust knowledge distillation}
To verify the advantages of the prototypes bridging the two feature spaces for global and local knowledge distillation, we conduct ablation experiments on the VOC dataset with $1\times$ learning schedule. ResNet101-based and ResNet50-based Faster R-CNN are used as the teacher and the student detectors, respectively.

We first separately apply $\mathcal{L}_{\rm global}$, $\mathcal{L}_{\rm local}^{\rm feat}$ and $\mathcal{L}_{\rm local}^{\rm resp}$ in Eq. \ref{eq-objective} for knowledge distillation. In our framework, the global knowledge is formed as the projections of instances w.r.t. the prototypes, while the feature-based and response-based local knowledge is weighted through the discrepancy of the projections. Table \ref{tb-ablation-prototype} shows that the prototypes can boost the global and local knowledge distillation by a large margin. By applying both the global and local knowledge, we show 1.6\% performance gain compared to the student detector with the $1\times$ learning schedule, which also surpasses the teacher detector with the mAP 82.4\%.

Furthermore, we also extend some existing methods based on the prototypes. DeFeat \cite{DecoupledFeature}, RKD \cite{RKD} and Vanilla-KD \cite{Hinton} are the representative feature-based, relation-based and response-based local knowledge distillation methods. We directly use the released source codes of FBKD and carefully re-implement the RKD (as RKD$^\dag$) and Vanilla-KD as (Vanilla-KD$^\dag$) for distilling the object detectors. Then, we apply the prototypes separately: as for RKD \cite{RKD}, we form the global knowledge by projecting the instances w.r.t. the prototypes; as for DeFeat \cite{DecoupledFeature} and Vanilla-KD \cite{Hinton}, we apply the distilling weight defined in Eq. \ref{eq-weight}, which is measured by the discrepancy w.r.t. the prototypes. Table \ref{tb-ablation-combination} shows the consistent performance gain among those three knowledge distillation methods with the prototypes, which shows the effectiveness of prototypes for both constructing more reliable global knowledge and more robust knowledge transfer. The implementation details by combining prototypes with those distilling methods are included in the supplemental materials.

\begin{table}[t]
\caption{Ablation experiment on separately applying the global knowledge $\mathcal{L}_{\rm global}$, feature-based local knowledge $\mathcal{L}_{\rm local}^{\rm feat}$ and response-based local knowledge $\mathcal{L}_{\rm local}^{\rm resp}$ in Eq. \ref{eq-objective} on VOC dataset with $1\times$ learning schedule.}
\label{tb-ablation-prototype}
\centering
\setlength{\tabcolsep}{3mm}
\begin{tabular}{c|c|ccccccc}
\toprule
\textbf{Module}&\textbf{Student}&\multicolumn{6}{c}{\textbf{Faster R-CNN Res101-50}} \\
\midrule
$\mathcal{L}_{\rm local}^{\rm feat}$ &  & \checkmark &  &  & \checkmark &  & \checkmark \\
$\mathcal{L}_{\rm global}$ &  &  & \checkmark &  &  & \checkmark & \checkmark \\
$\mathcal{L}_{\rm local}^{\rm resp}$ &   &   &   & \checkmark & \checkmark & \checkmark & \checkmark \\
\midrule
\textbf{AP}$_{50}$ & 81.3 & 82.0 & 82.4 & 82.2 & 82.6 & 82.4 & \textbf{82.9} \\
\bottomrule
\end{tabular}
\end{table}

\begin{table}[t]
\caption{Ablation experiment by combining the prototypes with the existing representative methods for feature-based \cite{DecoupledFeature}, relation-based \cite{RKD}, and response-based \cite{Hinton} local knowledge distillation, respectively.}
\label{tb-ablation-combination}
\centering
\setlength{\tabcolsep}{2.5mm}
\begin{tabular}{c|c|cc|cc|cc}
\toprule
\textbf{Method}&\textbf{Student}&\multicolumn{2}{c|}{\textbf{DeFeat} \cite{DecoupledFeature}} & \multicolumn{2}{c|}{\textbf{RKD}$^\dag$ \cite{RKD}} & \multicolumn{2}{c}{\textbf{Vanilla-KD}$^\dag$ \cite{Hinton}} \\
\midrule
\textbf{+prototypes} &  &  & \checkmark &  &   \checkmark &  &  \checkmark \\
\midrule
\textbf{AP}$_{50}$ & 81.3 & 82.0 & \textbf{82.4} & 81.6 & \textbf{82.0} & 81.8 & \textbf{82.2} \\
\bottomrule
\end{tabular}
\end{table}

\begin{table}[t]
    \caption{Ablation experiments on the hyperparameters $\alpha_1$, $\alpha_2$, $\alpha_3$, $\lambda$, $K$, and $T$.}
    \label{tb-ablation-hyper}
    \centering
\setlength{\tabcolsep}{2.5mm}
    \begin{tabular}{c|cccc||c|cccc}
   \toprule
    $\alpha_1$ & 0.1 & 0.5 & 1.0 & 1.2 & $\alpha_2$ & 0.5 & 0.8 & 1.0 & 1.5 \\
    \hhline{-----||-----}
    AP$_{50}$ & 82.1 & 82.3 & \textbf{82.9} & 82.6 & AP$_{50}$ & 82.3 & 82.4 & \textbf{82.9} & 82.6 \\
    \hline
    \hline 
    $\alpha_3$ & 1 & 5 & 10 & 20 & $\lambda$ & 1 & 10 & 50 & 100 \\
    \hline
    AP$_{50}$ & 82.3 & \textbf{82.9} & 82.3 & 82.2 & AP$_{50}$ & 82.3 & \textbf{82.9} & 82.6 & 82.0 \\
    \hline
    \hline
    $K$ & 1 & 5 & 10 & 20 & $T$ & 0.5 & 1 & 2 & 3 \\
    \hline
    AP$_{50}$ & 82.0 & 82.3 & \textbf{82.9} & 82.5 & AP$_{50}$ & 82.8 & \textbf{82.9} & 82.3 & 82.2 \\
    \bottomrule
    \end{tabular}
\end{table}

\subsection{Analysis of the hyperparameters}
We investigate the updating periods $T$ in Alg. \ref{alg-r2i} of the prototypes for knowledge distillation on the VOC dataset. Since the student detector is updated during training, the prototypes and their features $\bm{G}_s$ should be updated. Table \ref{tb-ablation-hyper} shows that as the period $T$ increasing, the performance slightly decreases. It is because the bootstrapped prototypes are approximations of the basis vectors of updated student detector, which results in some bias when forming the global knowledge as well as computing the discrepancy in Eq. \ref{eq-weight}. Table \ref{tb-ablation-hyper} also shows ablation experiments on the three weights $\alpha_1$, $\alpha_2$, and $\alpha_3$ of the three terms in Eq. \ref{eq-objective}, the number of selected prototypes $K$ and the similarity regularization weight $\lambda$ in Eq. \ref{eq-prototype-objective}. The results in Table \ref{tb-ablation-hyper} show that the proposed method is relatively robust to the hyperparameters, which achieves better performance than the student in a wide range of hyperparameters.

\begin{table}[t]
\caption{Ablation experiments on the effect of different prototype generation methods for knowledge distillation. For the cluster-like algorithms, e.g., K-Means and DBSCAN \cite{DBSCAN}, we apply them separately on the feature space of either the teacher or the student.}
\label{tb-ablation-pgm}
\centering
\setlength{\tabcolsep}{2mm}
\begin{tabular}{c|cc|cc|c|c}
\toprule
\textbf{Method}&\multicolumn{2}{c|}{\textbf{K-Means}} & \multicolumn{2}{c|}{\textbf{DBSCAN} \cite{DBSCAN}} & \textbf{Ambiguous} & \textbf{Ours} \\
\midrule
\textbf{Features} & Student & Teacher & Student & Teacher & - & Both \\
\textbf{AP}$_{50}$ & 82.3 & 82.1 & 82.4 & 82.2 & 81.8 & \textbf{82.9}  \\
\bottomrule
\end{tabular}
\end{table}

\begin{table}[b]
\caption{Knowledge distillation results with larger teacher on COCO dataset.}
\label{tb-large-coco}
\centering
\setlength{\tabcolsep}{1mm}
\begin{tabular}{c|c|ccc|ccc}
\toprule
\textbf{Method} & \textbf{Backbone} & \textbf{mAP} & \textbf{AP}$_{50}$ & \textbf{AP}$_{75}$ & \textbf{AP}$_s$ & \textbf{AP}$_m$ & \textbf{AP}$_l$\\
\midrule
Cascade R-CNN (teacher) & ResNext101 & 47.3 & 66.3 & 51.7 & 28.2 & 51.7 & 62.7 \\
Faster R-CNN (student) & ResNet50 & 38.4 & 59.0 & 42.0 & 21.5 & 42.1 & 50.3 \\
\midrule
FBKD \cite{FBKD} & ResNet50 & \textbf{41.5} & \textbf{62.2} & \textbf{45.1} & \textbf{23.}5 & 45.0 & 55.3 \\
Ours & ResNet50 & \textbf{41.5} & 61.9 & \textbf{45.1} & \textbf{23.5} & \textbf{45.1} & \textbf{55.4} \\
\bottomrule
\end{tabular}
\end{table}

\subsection{Analysis on the prototype generation methods}
In our framework, the prototypes play roles as the common basis vectors in both the feature spaces of the teacher and the student. They are selected by minimizing the reconstruction errors among instances in \emph{TS-space} as defined in Eq. \ref{eq-prototype-objective}. We also compare the proposed prototype generation algorithm in Alg. \ref{alg-pgm} with some other similar methods, e.g., the K-means and the DBSCAN \cite{DBSCAN}. Besides, we also deliberately select the same number of ambiguous instances, e.g., small or truncated instances, as the prototypes for comparison. In Table \ref{tb-ablation-pgm}, we show the performance of knowledge distillation based on those prototype generation methods. We find that the cluster-like algorithms, e.g., the K-means or the DBSCAN \cite{DBSCAN}, fail to improve the distillation by comparing the results in Table \ref{tb-ablation-prototype}, because those algorithms are applied only in the single feature space and can hardly bridge the two feature spaces of the teacher and the student. The poor performance by selecting the ambiguous instances as the prototypes further verify the importance of selecting the representative instances as the prototypes. Otherwise, it will bring large discrepancy as shown in Fig. \ref{fig-example} (right), and increase the difficulty of knowledge distillation.

\begin{figure}[t]
\centering
\includegraphics[width=12cm]{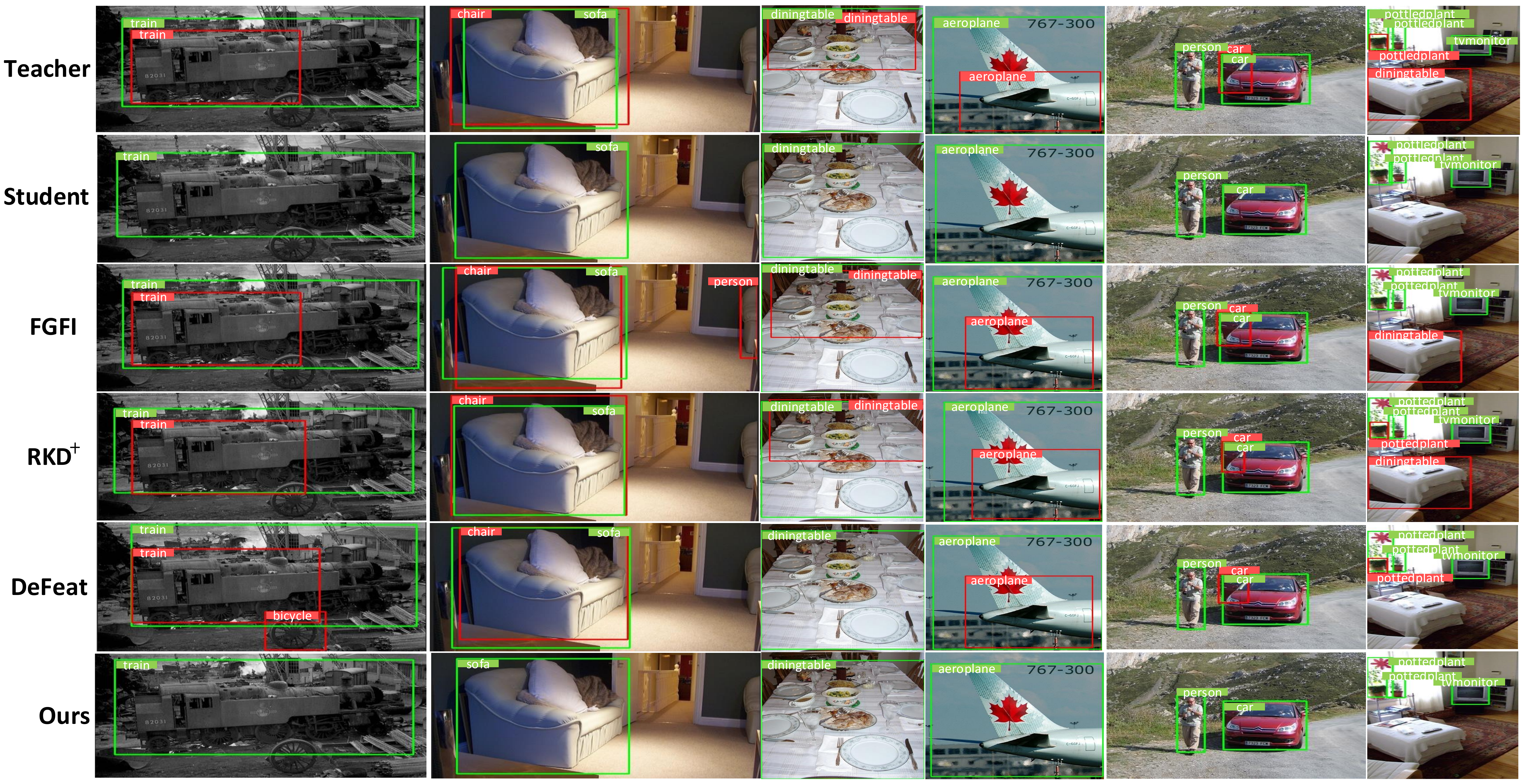}
\caption{Illustration of detection results via different knowledge distillation methods, e.g., FGIF \cite{FGFI}, our re-implemented RKD \cite{RKD}, DeFeat \cite{DecoupledFeature}, and ours. Some noisy knowledge of the teacher are transferred to the student (marked in red boxes). Best view in color.}
\label{fig-vis-det-results}
\end{figure}

\begin{table}[t]
\caption{Knowledge distillation results with larger teacher on VOC dataset.}
\label{tb-large-voc}
\centering
\setlength{\tabcolsep}{2mm}
\begin{tabular}{c|cc|ccc|c}
\toprule
\multirow{2}{*}{\textbf{Method}}&\multirow{2}{*}{\textbf{Teacher}}&\multirow{2}{*}{\textbf{Student}}&\multicolumn{4}{c}{\textbf{Faster R-CNN ResNet152-50}} \\
\cline{4-7} & & &  FGFI\cite{FGFI}&TADF\cite{Adaptive}&DeFeat\cite{DecoupledFeature}&Ours\\
\midrule
\textbf{AP}$_{50}$ & 83.1 & 81.3 & 81.6 & 81.7 & 82.3 & \textbf{82.9} \\
\bottomrule
\end{tabular}
\end{table}

\subsection{Distilling with larger teacher}
The larger teacher will achieve better performance, which might also bring an extra bonus for knowledge distillation. Following the common settings with the existing methods \cite{DecoupledFeature,FGFI,Adaptive}, on the VOC dataset, we use the Faster R-CNN with the backbones ResNet152 and ResNet50 as the teacher and the student. For a fair comparison, we follow DeFeat \cite{DecoupledFeature} by using 1$\times$ learning schedule. On the COCO dataset, we follow FBKD \cite{FBKD} by applying ResNeXt101-based \cite{ResNext} Cascade Mask R-CNN \cite{CascadeRCNN} as the teacher detector and the ResNet50-based Faster R-CNN as the student. The $2\times$ learning schedule is used as in FBKD \cite{FBKD}. In Table \ref{tb-large-voc} and Table \ref{tb-large-coco}, we show the performance of knowledge distillation with larger teachers on VOC and COCO datasets, respectively. The proposed method can still achieve the best performance on the VOC dataset, with the 0.6\% mAP 
advantage w.r.t. DeFeat \cite{DecoupledFeature}. On the COCO dataset, we achieve comparable performance as the FBKD \cite{FBKD} with the much larger teacher and heterogeneous backbone. The performance with larger teachers further shows the proposed method can be applied in various detection frameworks with the same hyperparameters.

\subsection{Analysis of noisy knowledge transferring}

In Figure \ref{fig-vis-det-results}, we also illustrate some wrong detection in red boxes, \emph{e.g.}, false positives and inaccurately located instances, from the teacher detector that are transferred to the student. Our method shows more promising results against noisy knowledge transferring and is capable to surpass the performance of the teacher detector. More quantitative analysis is discussed in the supplementary.

\section{Conclusion}
In this paper, we propose a novel knowledge distillation framework with global knowledge. The prototype generation module is first designed to find a group of common basis vectors, i.e., the \emph{prototypes}, by minimizing the reconstruction errors in both the feature spaces of the teacher and the student. The robust distillation module is then applied to (1) construct the global knowledge by projecting the instances w.r.t. the prototypes, and (2) robustly distill the global and local knowledge by measuring their discrepancy in the two spaces.
Experiments show that the proposed method achieves state-of-the-art performance on two popular detection frameworks and benchmarks. The extensive experimental results show that the proposed method can be easily stretched with larger teachers and the existing knowledge distillation methods to obtain further improvement.

\clearpage
\bibliographystyle{splncs04}
\bibliography{egbib}
\end{document}